\documentclass[letterpaper, 10 pt, conference]{ieeeconf}  % Comment this line out if you need a4paper

\IEEEoverridecommandlockouts                              % This command is only needed if 
\usepackage{amsmath} % assumes amsmath package installed
\usepackage{amssymb}  % assumes amsmath package installed
\usepackage{todonotes}
\let\labelindent\relax % deactivate ieee labelindent that conflicts with enumitem
\usepackage{enumitem}
\usepackage{mathtools}
\usepackage{rotating}
\usepackage{multirow}
\usepackage{siunitx}
\usepackage{caption}
\usepackage{subcaption}
\usepackage{float}
\usepackage{microtype}

\usepackage{booktabs} %nice table design

\usepackage{algorithmic} %to display algorithms
\usepackage{algorithm2e}
\graphicspath{{graphics/}}

\usepackage{tabto} % to use tabstops in lists

\usepackage{hyperref}
\hypersetup{
    colorlinks=true,
    linkcolor=black,
    filecolor=magenta,      
    urlcolor=blue,
    pdftitle={CarlaNCAP},
    pdfpagemode=None,
}
\usepackage{pifont}
\usepackage{booktabs} %nice table design
\usepackage{graphicx}

\title{\LARGE \bf
CarlaNCAP: A Framework for Quantifying the Safety of Vulnerable Road Users in Infrastructure-Assisted Collective Perception Using EuroNCAP Scenarios
}

\author{Jörg Gamerdinger$^{1}$, Sven Teufel$^{1}$, Simon Roller$^{1}$, and Oliver Bringmann$^{1}$
\thanks{$^{1}$University of T\"ubingen, Faculty of Science, Department of Computer Science, Embedded Systems Group 
\tt\small {\{joerg.gamerdinger, sven.teufel, simon.roller, oliver.bringmann\} @uni-tuebingen.de}}
}%

\begin{document}
\maketitle
\thispagestyle{empty}
\pagestyle{empty}

\begin{abstract}
The growing number of road users has significantly increased the risk of accidents in recent years. Vulnerable Road Users (VRUs) are particularly at risk, especially in urban environments where they are often occluded by parked vehicles or buildings. Autonomous Driving (AD) and Collective Perception (CP) are promising solutions to mitigate these risks. In particular, infrastructure-assisted CP, where sensor units are mounted on infrastructure elements such as traffic lights or lamp posts, can help overcome perceptual limitations by providing enhanced points of view, which significantly reduces occlusions. To encourage decision makers to adopt this technology, comprehensive studies and datasets demonstrating safety improvements for VRUs are essential. In this paper, we propose a framework for evaluating the safety improvement by infrastructure-based CP specifically targeted at VRUs including a dataset with safety-critical EuroNCAP scenarios (CarlaNCAP) with 11k frames. Using this dataset, we conduct an in-depth simulation study and demonstrate that infrastructure-assisted CP can significantly reduce accident rates in safety-critical scenarios, achieving up to 100\,\% accident avoidance compared to a vehicle equipped with sensors with only 33\,\%.
Code is available at \url{https://github.com/ekut-es/carla_ncap}
\end{abstract}

%\begin{IEEEkeywords}
%    Lane Detection, Safety Evaluation, Autonomous Driving
%\end{IEEEkeywords}

%%%%%%%%%%%%%%%%%%%%%%%%%%%%%%%%%%%%%%%%%%%%%%%%%%%%%%%%%%%%%%%%%%%%%%%%%%%%%%%%
\section{INTRODUCTION}
\label{sec:intro}
Human error is responsible for \SI{90}{\percent} of fatal road accidents in the European Union~\cite{EuropeanUnion2019}. Autonomous driving is expected to increase road safety and improve traffic flow by eliminating human error. To achieve this, automated vehicles need to overcome a number of challenges before they are ready to enter the market. Accurate and complete perception of the environment is one of the most important issues. Unfortunately, perception is not always accurate, increasing the risk of fatal accidents involving autonomous vehicles.  
Collective Perception (CP) is a promising approach to overcome the limitations of  vehicle-local perception. In particular, infrastructure-assisted collective perception (IACP) leads to an increase in perception performance~\cite{schiegg_infrastructure,10562184,vijay2021optimal}.
Using comprehensive studies make an important contribution to demonstrating the need for infrastructure sensor deployment and may also contribute to better societal acceptance of autonomous systems.
As shown by Volk et al.~\cite{volk2020safety}, it is not sufficient to consider only performance metrics when evaluating perception systems. Additional factors should be considered to determine the safety of a perception system. Additionally, to determine safety, safety-critical scenarios must be considered.

In the field of infrastructure-based CP, several datasets exist such as DOLPHINS~\cite{dolphins}, Deep-Accident~\cite{wang2023deepaccident}, TUMTraf-V2X~\cite{zimmer2024tumtraf} and SCOPE~\cite{SCOPE-Dataset}. However, none of these datasets includes safety-critical situations with vulnerable road users (VRUs) such as pedestrians or cyclists who are particularly worthy of protection. Therefore, we present a novel framework for investigating infrastructure-assisted CP to improve the safety of VRUs using safety-critical EuroNCAP scenarios. 

Our main contributions are:
\begin{itemize}
    \item We present a novel framework to investigate the safety of VRUs in urban environments.
    \item We propose a synthetic dataset with EuroNCAP scenarios to test VRU perception in urban environments.
    \item We quantify the effect of various infrastructure sensor setups for VRU safety enhancement.
\end{itemize}

In Sec.~\ref{sec:related_work} we give an overview on previous studies identifying the improvement of perception capabilities by infrastructure-assisted collective perception and related datasets. After that, we present a novel dataset including safety-critical EuroNCAP scenarios which improves the state-of-the-art to investigate the safety enhancement for VRUs. In Sec.~\ref{sec:experiments}, we discuss the conducted VRU safety enhancement study and the results. Finally, we conclude our work and give an outlook to further research. 

%%%%%%%%%%%%%%%%%%%%%%%%%%%%%%%%%%%%%%%%%%%%%%%%%%%%%%%%%%%%%%%%%%%%%%%%%%%%%%%%
\section{RELATED WORK}
\label{sec:related_work}

    Schiegg et al.~\cite{schiegg_infrastructure} presented a simulation study on the impact of IACP for different V2X equipment rates. They proposed optimization strategies to take into account the special role of infrastructure sensors. Besides the reduction of communication loss due to packet duplication, they demonstrated the increase of perception performance using infrastructure cameras in a motorway scenario. They showed that the perception performance in terms of average accuracy can be increased by up to \SI{100}{\percent} compared to vehicle-local perception using adapted message generation rules. For the standard ETSI message generation rules, an increase of about \SI{50}{\percent} could be demonstrated. However, they only considered a motorway scenario; hence, no VRUs are included. Furthermore, the dataset used is not publicly available.
    Ahmad et al.~\cite{10562184} introduced the concept of Cooperative Infrastructure Perception. It fuses outputs from multiple infrastructure sensors to generate accurate and timely perception outputs. This approach overcomes the line-of-sight limitations of vehicle-mounted sensors and has shown potential for improving vehicle safety and traffic throughput at intersections. However, the dataset used is recorded at a single intersection and therefore does not provide much variety, nor does it include safety-critical situations.
    The topic of sensor placement for infrastructure-assisted CP has been investigated in~\cite{vijay2021optimal, du2022quantifying, chen2024roadside}. Vijay et al.~\cite{vijay2021optimal} proposed a framework that combines simulation and optimization techniques to determine the most cost-effective sensor placement in terms of maximizing sensor coverage and minimizing blind spots. The approach of Du et al.~\cite{du2022quantifying} uses an occlusion degree model to investigate the impact of sensor configurations on detection performance. However, none of these works consider VRU safety and use a dataset with safety-critical situations. Chen et al.~\cite{chen2024roadside} also propose a stochastic method to optimize the placement of infrastructure LiDAR sensors. They used a motorway scenario to determine the effect of different sensor positions. However, their study does not include any safety-critical situations and there are no VRUs. All the research pointed out that the performance of infrastructural sensor units is highly dependent on their placement, but lacks evaluation on safety-critical scenarios including VRUs. For further methods on the placement of sensors we refer to Chen et al.~\cite{chen2024roadside}.

    As presented by Teufel et al.~\cite{dataset-review}, there are several collective perception datasets including roadside units (RSUs).
    In 2022, Yu et al.~\cite{yu2022dair} presented the DAIR-V2X dataset, which was the first large-scale collective perception dataset with real-world data. It was recorded using roadside sensors at several intersections and a CAV. The data is recorded at \SI{10}{\hertz} using two 300-layer \SI{100}{\degree} LiDAR sensors with a range of \SI{280}{\metre} and two 1920$\times$1080\,px RGB cameras for the RSUs and a 40-layer \SI{360}{\degree} LiDAR and a 1920$\times$1080\,px RGB camera for the CAV. In total, the dataset consists of 13,000 frames, including different types of vehicles as well as VRUs (pedestrians and cyclists). The V2X-Seq dataset by Yu et al.\cite{v2xseq} is an extension of the DAIR-V2X dataset with 15k frames. The dataset uses the same sensors and provides trajectories. However, both datasets are not publicly available and therefore cannot be used to evaluate safety enhancement for VRUs. Also in 2022 Xu et al.~\cite{xu2022v2xvit} presented V2XSet. The synthetic dataset contains a total of 11,447 frames which are generated using the CARLA simulator~\cite{carla}. The dataset features a wide range of scenarios and a varying number of CAVs equipped with a 32-layer \SI{360}{\degree} LiDAR sensor and four RGB cameras (800$\times$600\,px). Some of the scenarios include one roadside unit with the same sensors. However, the dataset contains different anomalies such as unrealistic traffic and also lacks VRUs. The DOLPHINS dataset by Mao et al.~\cite{dolphins} is a synthetic dataset generated with CARLA. The dataset has 42,376 frames with 292,549 objects at six different scenarios. Some of the scenarios contain one or two RSUs besides an ego vehicle. The vehicles as well as the RSUs are equipped with a RGB camera (1920$\times$1080\,px) and a 64-layer \SI{360}{\degree} LiDAR sensor with a sensing range of \SI{200}{\metre}. Two scenarios contains weather as well as VRUs. However, the data frequency of \SI{2}{\hertz} is rather low, thus, the dataset is not suitable to evaluate for safety-critical tasks. Additionally, the dataset is not publicly available. The synthetic V2X-Sim2.0 dataset~\cite{V2XSim} includes CAVs and RSUs. 2-5 CAVs per scenario are equipped with a 32-layer LiDAR and 6 RGB cameras (1600$\times$900\,px), the RSUs are equipped with a 32-layer LiDAR and 4 RGB cameras (1600$\times$900\,px). The recordings result in a total of 10,000 frames. As passive traffic, the dataset includes various classes of vehicles and motorcyclists. However, pedestrians and cyclists are not included and the number of frames is rather low. The dataset is therefore not suitable for evaluating the safety of VRUs. A dataset that includes safety-critical scenarios is the synthetic DeepAccident dataset by Wang et al.~\cite{wang2023deepaccident}. This dataset contains different accident scenarios with a total 57k frames. In each scenario, four CAVs and one RSU are equipped with a 32-layer \SI{360}{\degree} LiDAR sensor with \SI{70}{\metre} range, recording at \SI{10}{\hertz} and 6 RGB cameras with 1600$\times$900\,px and \SI{70}{\degree} FoV. The dataset includes VRUs, but they are not part of the collision scenarios, making the dataset unsuitable for evaluating VRU safety. The V2X-Real real-world dataset was presented by Xiang et al. \cite{xiang2024v2x}. Two CAVs and two RSUs equipped with 128- and 64-layer \SI{360}{\degree} LiDAR sensors and 2-4 1920$\times$1080\,px RGB cameras are used for recording. The 33k frames contain objects from 10 classes, including various vehicles, pedestrians and cyclists. However, no safety-critical scenarios are included, making it unsuitable for determining safety. Zimmer et al.~\cite{zimmer2024tumtraf} presented the real-world TUMTraf V2X Collective Perception dataset recorded at a large intersection using an RSU and a CAV. The RSU was equipped with a 64-layer \SI{360}{\degree} LiDAR with \SI{120}{\metre} range and four 1920$\times$1200\,px RGB cameras. The CAV was equipped with a 32-layer \SI{360}{\degree} LiDAR with a \SI{200}{\metre} range and a 1920$\times$1200\,px RGB camera. The dataset consists of 1000 frames. This small number makes the dataset unsuitable for training and evaluation of perception systems. In addition, the number of RSUs is too small to determine a good setup. The SCOPE dataset by Gamerdinger et al.~\cite{SCOPE-Dataset} is a synthetic dataset generated using CARLA simulator. The dataset consists of 17,600 frames from 44 different scenarios with varying weather conditions. The scenarios contain a varying number of CAVs depending on the scenario type. Reasonable scenarios are also equipped with up to 4 RSUs. The CAVs and RSUs are equipped with two different \SI{360}{\degree} and one Solid-State LiDAR and 5 cameras with 1920$\times$1080\,px Additionally, they used an improved LiDAR model and two additional maps created by their own to achieve a more realistic and diverse dataset. SCOPE also includes a wide range of scenarios including VRUs. However, the main goal for this dataset is a realistic evaluation of collective perception and represent realistic traffic but not safety-critical scenarios.
\begin{figure*}[t!]
    \centering
    \begin{subfigure}[t]{0.32\textwidth}
        \centering
        \includegraphics[height=1.4in, trim={3cm 0.2cm 3cm 0cm}]{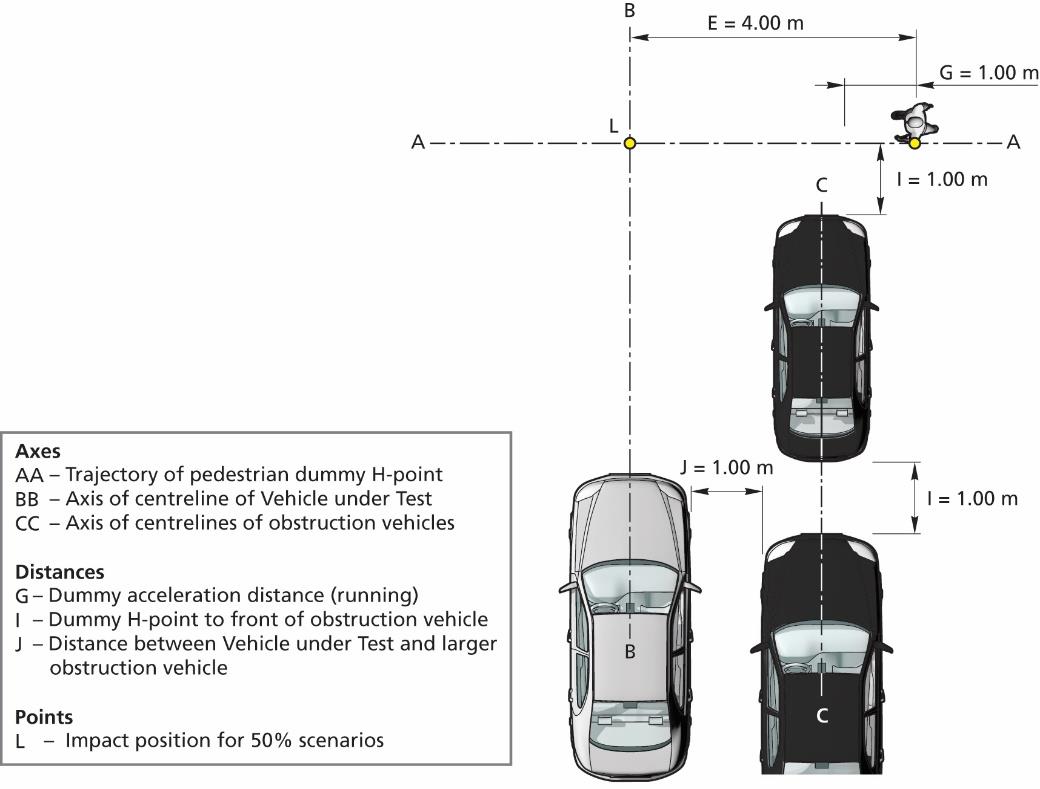}
        \caption{CPNC50}
        \label{fig:cpnc}
    \end{subfigure}%
    \begin{subfigure}[t]{0.32\textwidth}
        \centering
        \includegraphics[height=1.3in, trim={3cm 0.2cm 3cm 0cm}]{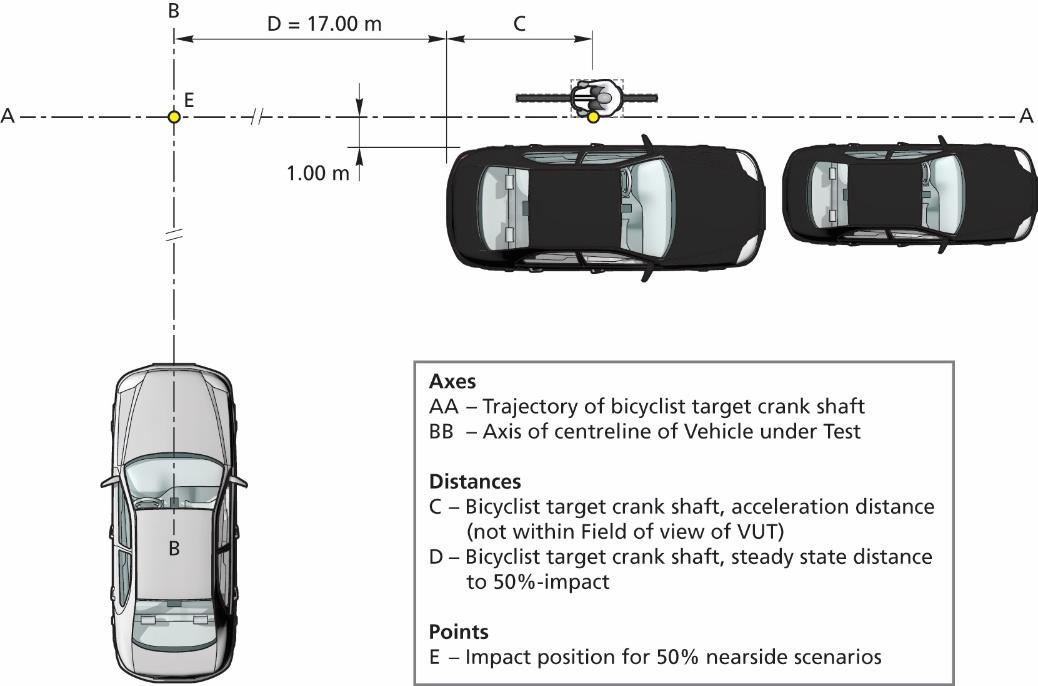}
        \caption{CBNA}
        \label{fig:cbna}
    \end{subfigure}
    \begin{subfigure}[t]{0.32\textwidth}
        \centering
        \includegraphics[height=1.4in, trim={3cm 0.2cm 3cm 0cm}]{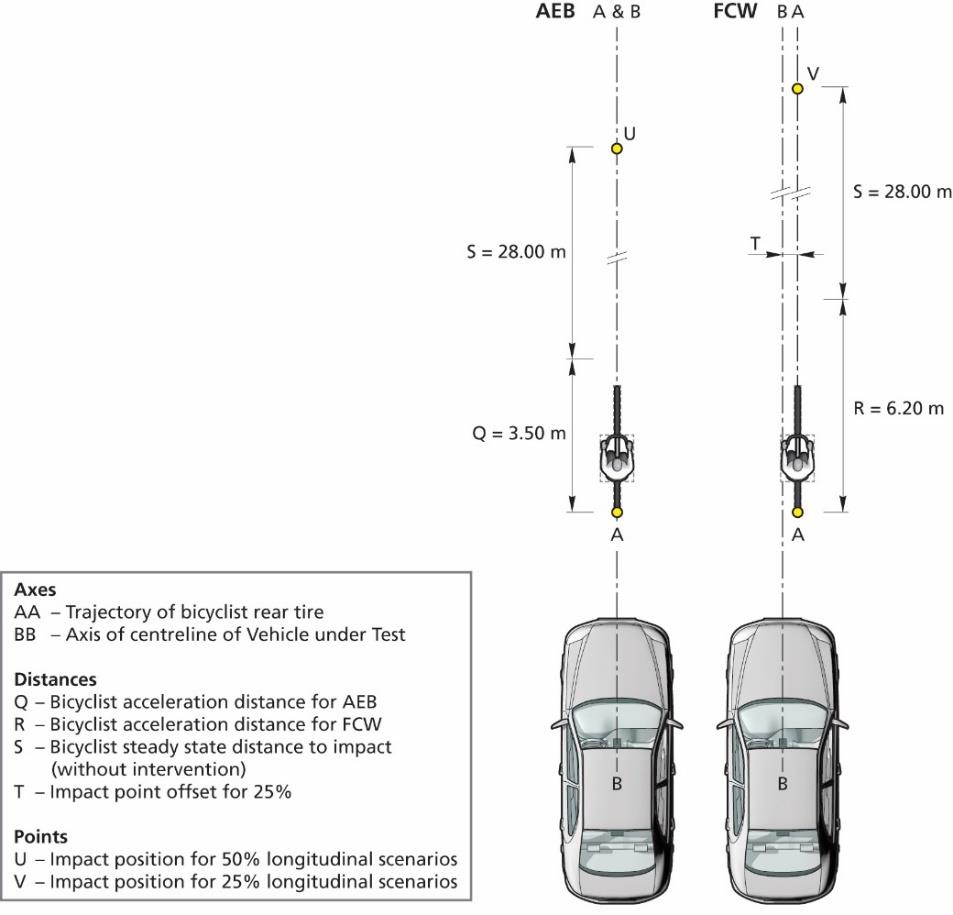}
        \caption{CBLA}
        \label{fig:cbla}
    \end{subfigure}
    \vspace*{5mm}
    
    \begin{subfigure}[t]{0.32\textwidth}
        \centering
        \includegraphics[height=1.4in, trim={3cm 0.2cm 3cm 0cm}]{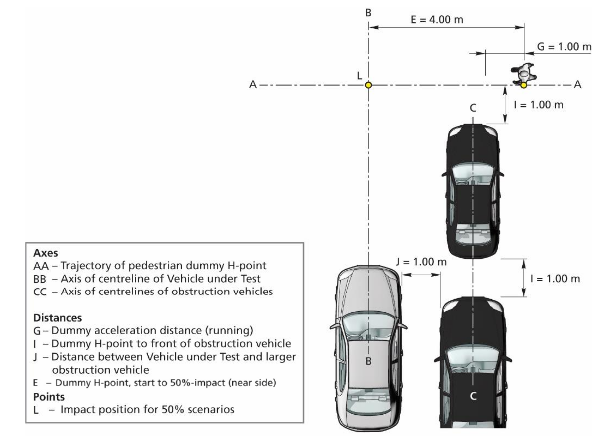}
        \caption{CPNCO}
        \label{fig:cpnco}
    \end{subfigure}%
    \begin{subfigure}[t]{0.32\textwidth}
        \centering
        \includegraphics[height=1.3in, trim={3cm 0.2cm 3cm 0cm}]{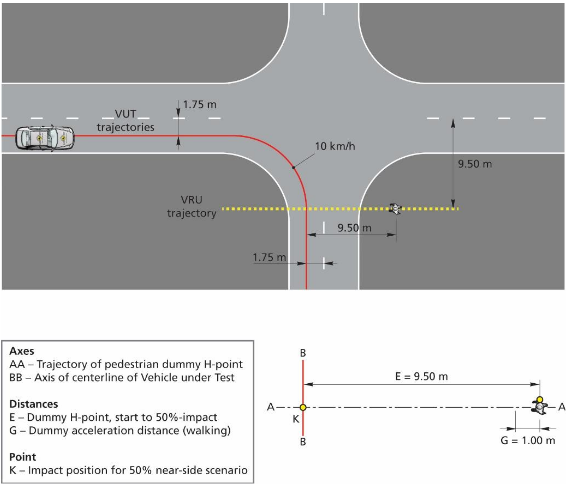}
        \caption{CPTA}
        \label{fig:cpta}
    \end{subfigure}
    \begin{subfigure}[t]{0.32\textwidth}
        \centering
        \includegraphics[height=1.4in, trim={3cm 0.2cm 3cm 0cm}]{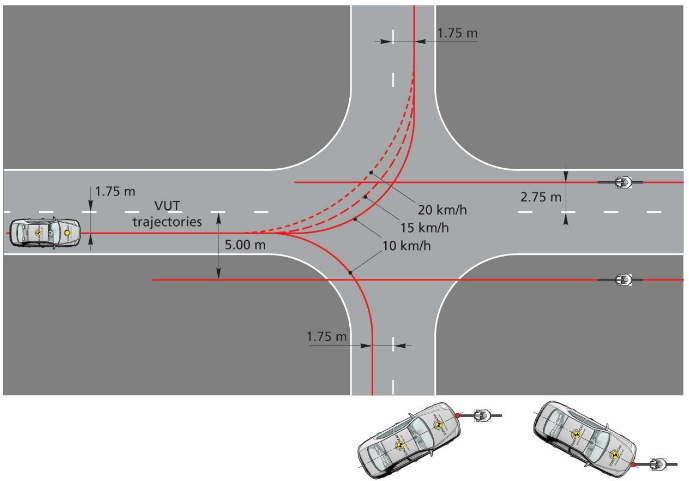}
        \caption{CBTA}
        \label{fig:cbta}
    \end{subfigure}
    \caption{EuroNCAP VRU AEB scenarios for the CARLA-NCAP dataset. Figures from~\cite{EuroNCAP}}
    \label{fig:scenarios}
\end{figure*}
A framework for CARLA-based pedestrian dataset generation was presented by Wielgosz et al.~\cite{wielgosz2023carlabspsimulateddatasetpedestrians}. The framework is highly configurable and the resulting dataset consists of about 225k frames. However, the goal of the dataset is the prediction of pedestrian crossings. The dataset therefore contains only recordings from one RGB camera with 1600$\times$600\,px and a single pedestrian walking on a sidewalk or crossing the street. As a result, the dataset does not contain a reasonable sensor setup to evaluate infrastructure-assisted CP, nor does it contain safety-critical situations.

\section{CARLANCAP DATASET}
\label{sec:dataset}

In order to evaluate the safety enhancement for VRUs due to infrastructure-assisted CP, an appropriate dataset is required. This dataset must be appropriate in terms of defined scenarios which are safety-critical and lead to a collision without a detection of the VRU. However, none of the presented datasets from Sec.~\ref{sec:related_work} includes such safety-critical scenarios with VRUs. Moreover, simple and defined scenarios with varying velocities are helpful for a better comparison and interpretability of the results. In this context, the EuroNCAP Automated Emergency Braking (AEB) protocol~\cite{EuroNCAP} is particularly well-suited, as it delineates a range of test cases for pedestrians and cyclists of varying velocities. Due to safety concerns conducting such tests in real-world settings is not recommended. Additionally, the utilization of real-world test dummies often exhibits inadequate realistic movement, a limitation that can compromise the interpretability of the results. In light of these considerations, we put forward a novel synthetic NCAP dataset (CarlaNCAP) that has been recorded using the CARLA simulator~\cite{carla}, which incorporates camera and LiDAR data from a Vehicle-under-Test (VUT), along with 12 roadside units, with the aim of ascertaining the impact of infrastructural-sensor units on the safety of VRUs. The dataset features three EuroNCAP scenarios (see Sec.~\ref{subsec:scenarios}) with varying VUT speeds as proposed by the EuroNCAP AEB VRU test protocol~\cite{EuroNCAP}.
%The dataset contains 11,134 frames, giving rise to a total of 144,742 images and LiDAR point clouds.

\subsection{Scenarios}
\label{subsec:scenarios}
The scenarios from the EuroNCAP AEB Less VRU standard are shown in Fig.~\ref{fig:scenarios}. There is one scenario including a pedestrian (CPNC-50) and two scenarios with cyclists (CBNA and CBLA). 
An overview of the required speeds for the VUT and the VRU is shown in Tab.~\ref{tab:scenario_speeds}. The speed for the vehicle is increased by \SI{5}{\kilo\metre\per\hour} in each iteration.

\begin{table}[h!]
    \centering
    \caption{VUT and VRU speed for EuroNCAP AEB scenarios, based on~\cite{EuroNCAP}}
    \begin{tabular}{c    c    c} \toprule
    Scenario & VUT Speed & VRU Speed  \\ \midrule
    CPNC-50 / CPNCO  & \SIrange{20}{60}{\kilo\metre\per\hour}  &\SI{5}{\kilo\metre\per\hour}  \\
    CPLA  & \SIrange{20}{60}{\kilo\metre\per\hour}  &\SI{5}{\kilo\metre\per\hour}  \\
    CPTA & \SIrange{10}{20}{\kilo\metre\per\hour}  &\SI{5}{\kilo\metre\per\hour}  \\
    CBNA  & \SIrange{20}{60}{\kilo\metre\per\hour}  &\SI{15}{\kilo\metre\per\hour}  \\
    CBLA  & \SIrange{25}{60}{\kilo\metre\per\hour}  &\SI{15}{\kilo\metre\per\hour}  \\
    CBLA  & \SIrange{10}{20}{\kilo\metre\per\hour}  &\SI{15}{\kilo\metre\per\hour}  \\
 \bottomrule
    \end{tabular}
    \label{tab:scenario_speeds}
    \vspace*{-0.2cm}
\end{table}

\paragraph{CPNC-50}
The CPNC-50 scenario is a pedestrian scenario where the pedestrian is crossing the VUT's path from the right side (see Fig.~\ref{fig:cpnc}). While moving towards the VUT trajectory, the pedestrian is occluded by parked vehicles which makes it difficult to perceive. The VUT speed varies between \SIrange{20}{60}{\kilo\metre\per\hour} while the pedestrian moves with \SI{5}{\kilo\metre\per\hour}.

\paragraph{CPNCO}
The CPNCO scenario is a pedestrian scenario where the pedestrian is crossing the VUT's path from the right side (see Fig.~\ref{fig:cpnco}) such as for the CPNC-50 scenario. However, the pedestrian in this scenario is a child which makes it more difficult to perceive due to the reduced height. The speeds are equal to the CPNC-50 scenario.

\paragraph{CPLA}
The CPLA scenario in which the pedestrian walks in front of the VUT with \SI{5}{\kilo\metre\per\hour} while the VUT speed ranges from \SIrange{20}{60}{\kilo\metre\per\hour}. The vehicle must brake to avoid a collision. In this scenario the relative angle of the pedestrian to the VUT leads to a very narrow silhouette which makes the detection more difficult.

\paragraph{CPTA left/right turn}
The last pedestrian scenario is the CPTA scenario in which the VUT is turning to the left or right (see Fig.~\ref{fig:cpta}) and a pedestrian is crossing the street and the VUT's trajectory. This is a very likely scenario in urban environments and due to the relative angle the VRU is within a potential blind spot. The VUT speed ranges between \SIrange{10}{20}{\kilo\metre\per\hour} for left turn and \SI{10}{\kilo\metre\per\hour} for the right turn. The pedestrian is moving with \SI{5}{\kilo\metre\per\hour}.  

\paragraph{CBNA}
In the CBNC scenario, the VRU is represented by a cyclist moving from the right crossing the VUT path (see Fig.~\ref{fig:cbna}), with a speed of \SI{15}{\kilo\metre\per\hour}. The speed of the VUT ranges from \SIrange{20}{60}{\kilo\metre\per\hour}. The cyclist is occluded by an object until \SI{17}{\metre} before the trajectories intersection point.

\paragraph{CBLA}
The CBLA scenario in which the cyclist drives in front of the VUT with \SI{15}{\kilo\metre\per\hour} while the VUT speed ranges from \SIrange{25}{60}{\kilo\metre\per\hour}. The vehicle must brake to avoid a collision. In this scenario the relative angle of the cyclist to the VUT leads to a very narrow silhouette which makes the detection more difficult.

\paragraph{CBTA}
The last scenario is the CPTA scenario in which the VUT is turning to the left or right (see Fig.~\ref{fig:cbta}) and a cyclist is crossing the street and the VUT's trajectory. This is a very likely scenario in urban environments and due to the relative angle the VRU is within a potential blind spot. The VUT speed ranges between \SIrange{10}{20}{\kilo\metre\per\hour} for left turn and \SI{10}{\kilo\metre\per\hour} for the right turn. The cyclist is moving with \SI{15}{\kilo\metre\per\hour}.  

\begin{figure}[b]
    \centering
    \includegraphics[width=\linewidth]{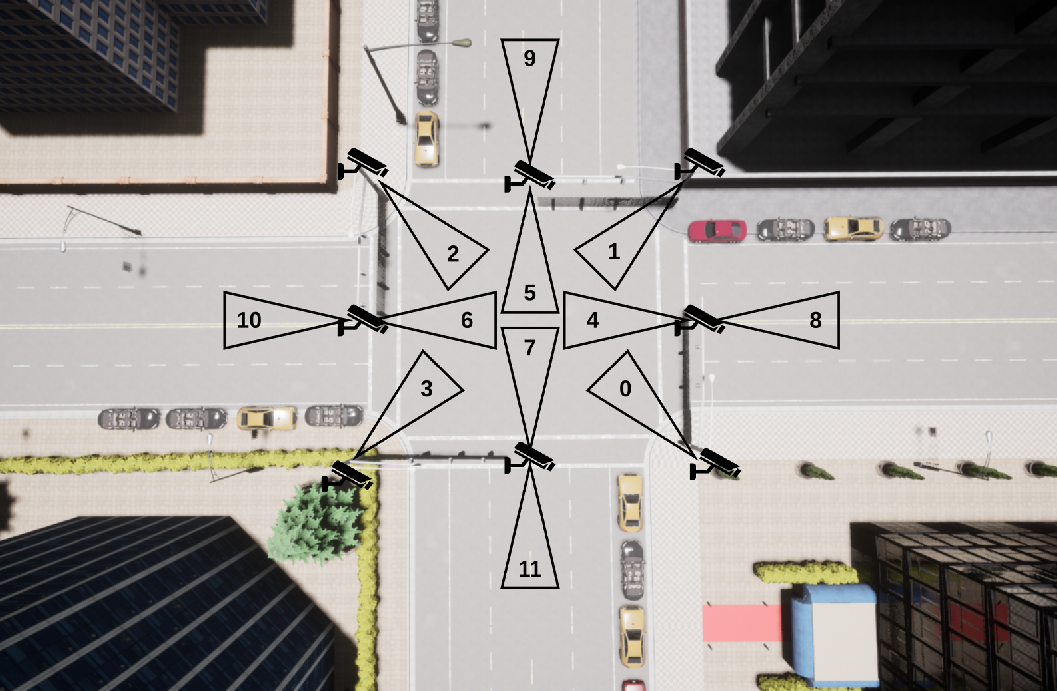}
    \caption{Symbolic representation of the camera placement for the test intersection of the CARLA-NCAP dataset}
    \label{fig:rsu_placement}
\end{figure}
\subsection{Sensor Setup}
For each scenario one vehicle and 12 RSUs are equipped with sensors. Each sensor unit consists of one RGB camera and one semantic segmentation camera with a resolution of 1920$\times$1080\,px and \SI{90}{\degree} Field-of-View (FoV).
As camera sensors have some disadvantages during night time or under adverse weather, we additionally equip the VUT and the RSUs with a Solid-State-LiDAR sensor. As the VRU is always in front of the VUT and the RSU sensors are tilted down towards the street it is not required to use a \SI{360}{\degree} LiDAR. The recording frequency of all sensors is \SI{10}{\hertz}. More information about the sensor specification can be found in Tab.~\ref{tab:sensor_spec}.

\begin{figure*}[t!]
    \centering
    \begin{subfigure}[t]{0.32\textwidth}
        \centering
        \includegraphics[width=\textwidth]{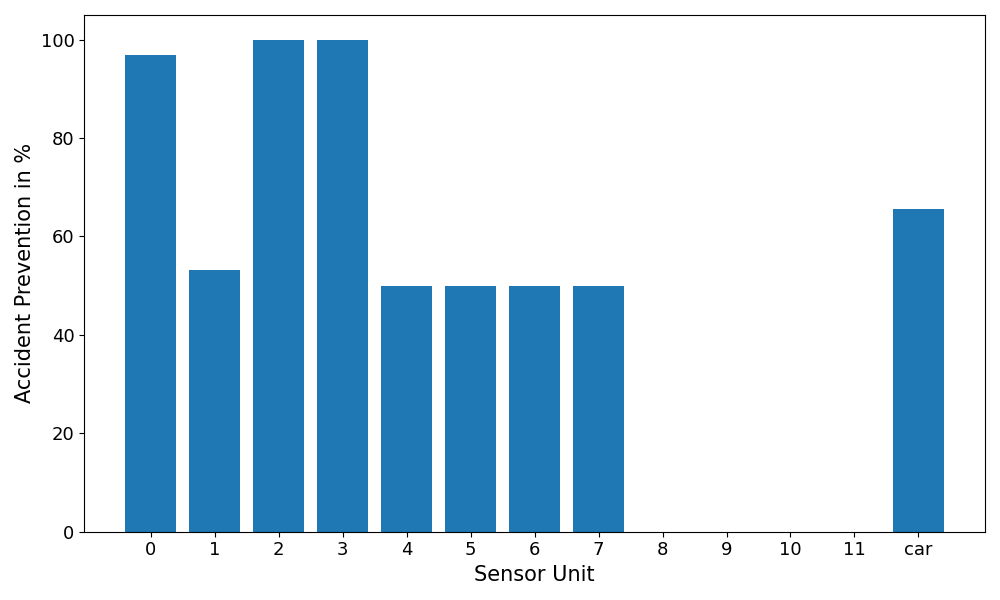}
        \caption{CPNC50}
        \label{fig:results_safety_cpnc}
    \end{subfigure}%
    \begin{subfigure}[t]{0.32\textwidth}
        \centering
        \includegraphics[width=\textwidth]{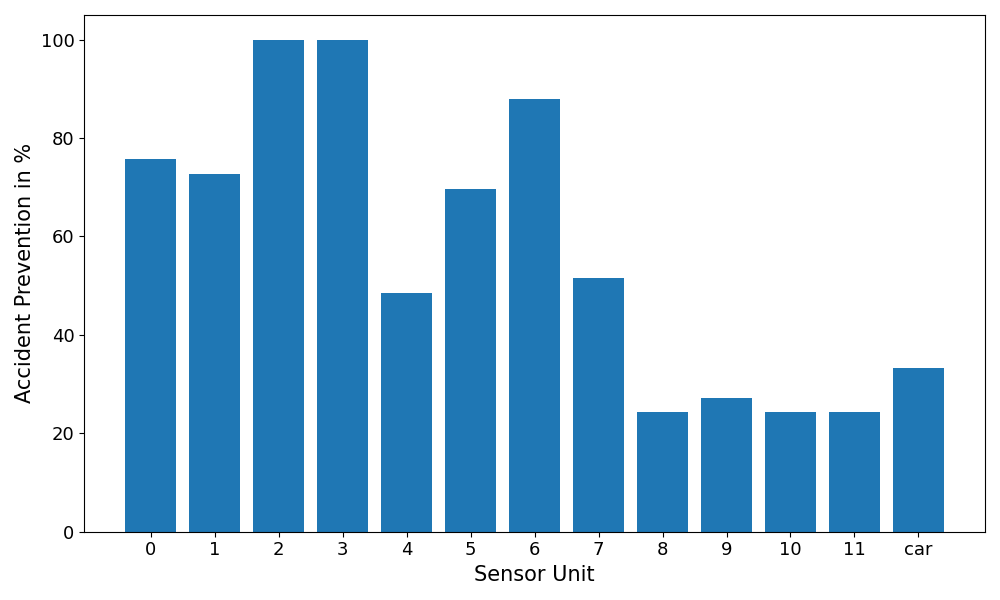}
        \caption{CBNA}
        \label{fig:results_safety_cbna}
    \end{subfigure}
    \begin{subfigure}[t]{0.32\textwidth}
        \centering
        \includegraphics[width=\textwidth]{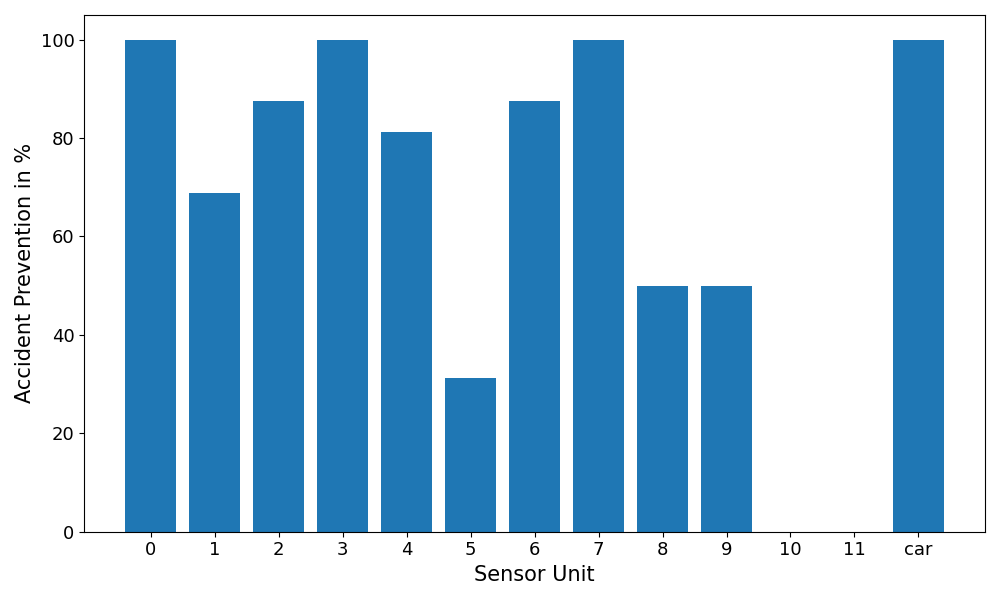}
        \caption{CBLA}
        \label{fig:results_safety_cbla}
    \end{subfigure}
    \caption{Results of the analysis of accident prevention capabilities on the EuroNCAP VRU AEB scenarios CPNC-50 (a), CBNA (b) and CBLA(c).}
    \label{fig:results_safety}
\end{figure*}
\begin{table}[t]
\centering
\caption{Sensor specification}
\begin{tabular}{l  l} \toprule
    Sensor & Specifications \\ \midrule
    1$\times$ RGB Camera  & 1920$\times$1080\,px, \SI{110}{\degree} FOV, \SI{10}{\hertz} \\
    \rule{0pt}{3ex}1$\times$ SemSeg Camera & 1920$\times$1080\,px, \SI{110}{\degree} FOV ,\SI{10}{\hertz}\\
    \rule{0pt}{3ex}1$\times$ CUBE LiDAR & \SI{72}{\degree}$\times$\SI{30}{\degree} FOV, 52 Scan Lines, \SI{10}{\hertz},\\& \SI{250}{\metre} range, 90k points per \si{\second}\\ \bottomrule
\end{tabular}
\vspace{-1mm}
\label{tab:sensor_spec}
\end{table}

The RSU sensors are positioned at reasonable positions around the test intersection in a height of about \SI{7}{\metre} at lamp posts and traffic lights. The positioning of the RSU sensors is shown in Fig.~\ref{fig:rsu_placement}. 

%%%%%%%%%%%%%%%%%%%%%%%%%%%%%%%%%%%%%%%%%%%%%%%%%%%%%%%%%%%%%%%%%%%%%%%%%%%%%%%%

\section{VRU Safety Study}
\label{sec:experiments}

In order to evaluate the enhancement of VRU safety using infrastructure-assisted CP, we use the dataset presented in Sec.~\ref{sec:dataset}. The scenarios are executed on a intersection of CARLA \textit{Town05} with each scenario being performed from different incoming directions in order to achieve greater diversity. For the object detection, the camera-based Faster R-CNN detector~\cite{ren2016faster} is employed. A detection of the VRU is considered as correct if the Intersection over Union (IoU) is higher than 0.5. As communication overhead and processing time, we use \SI{25}{\milli\second} as suggested by Volk et al.~\cite{volk2021}. For the emergency braking maneuver to avoid a collision, the braking acceleration ($a=$\SI{7.72}{\metre\per\second\squared}) from the EuroNCAP protocol is utilized.

\subsection{Metrics}
\label{subsec:metrics}
For the evaluation we want to focus on safety.In terms of safety we consider the collision avoidance rate. A collision is counted as avoided if the perception of the VRU is early enough to perform an emergency braking maneuver of the VUT which avoids the collision. To avoid false alarms, the VRU must be perceived for at least three consecutive frames. 

\subsection{Results}
\label{subsec:results}

\begin{figure*}[t]
    \centering
    \begin{subfigure}[t]{0.43\textwidth}
        \centering
        \includegraphics[width=\linewidth]{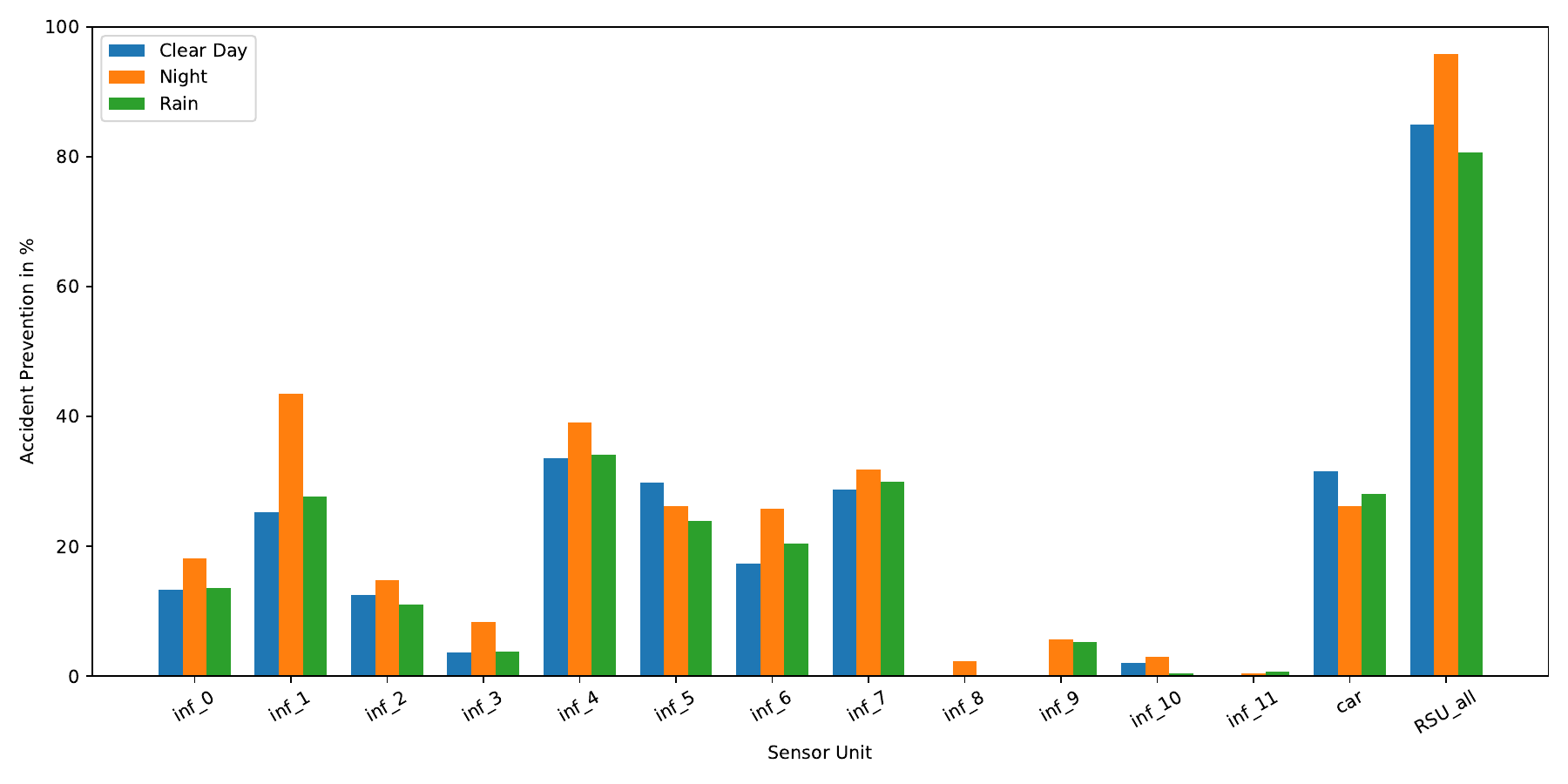}
        \caption{CBNA}
        \label{fig:CBNA_results}
    \end{subfigure}%
    \begin{subfigure}[t]{0.43\textwidth}
        \centering
        \includegraphics[width=\linewidth]{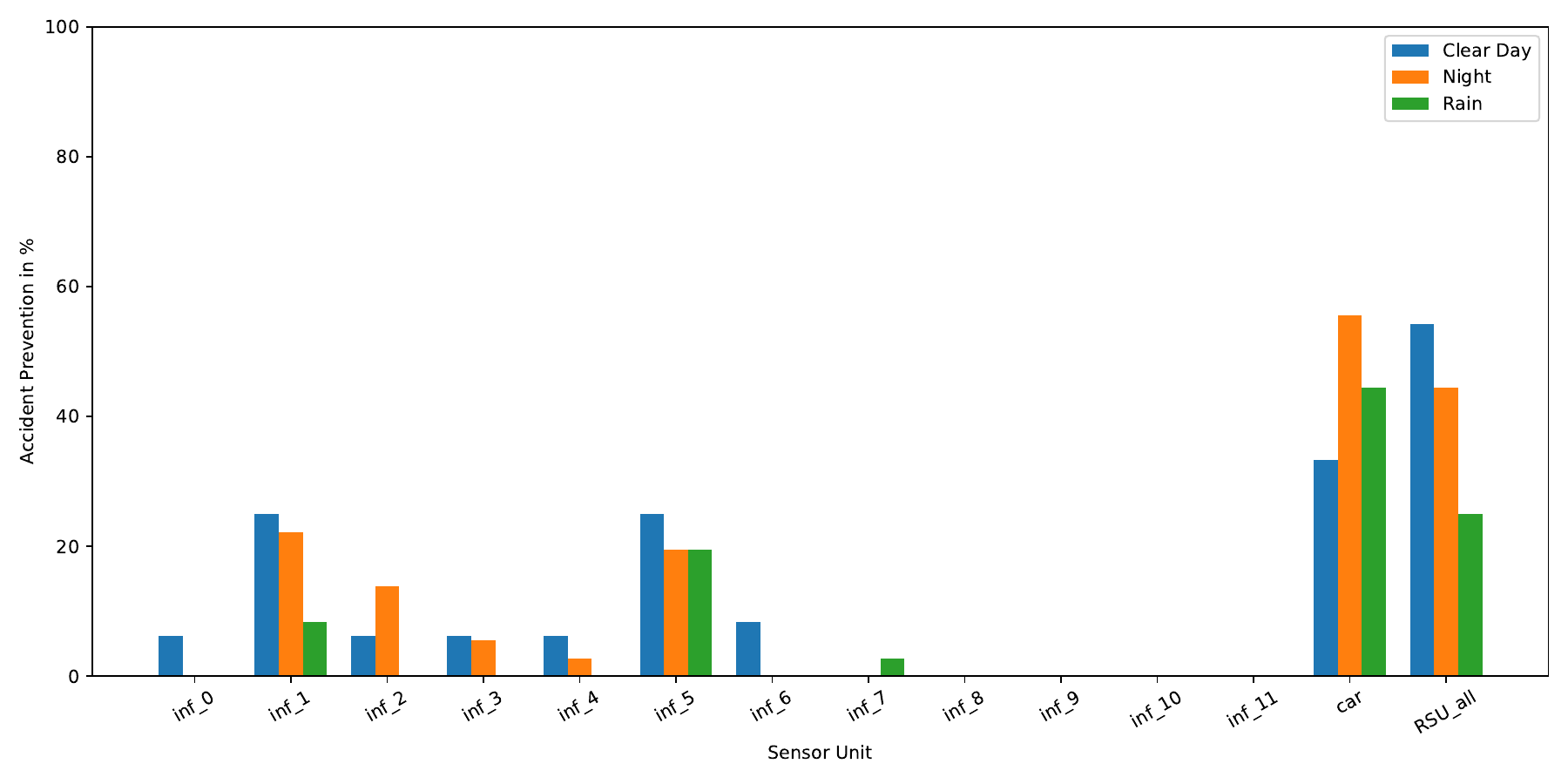}
        \caption{CBTAno}
        \label{fig:CBTAno_results}
    \end{subfigure}
    
    \begin{subfigure}[t]{0.43\textwidth}
        \centering
        \includegraphics[width=\linewidth]{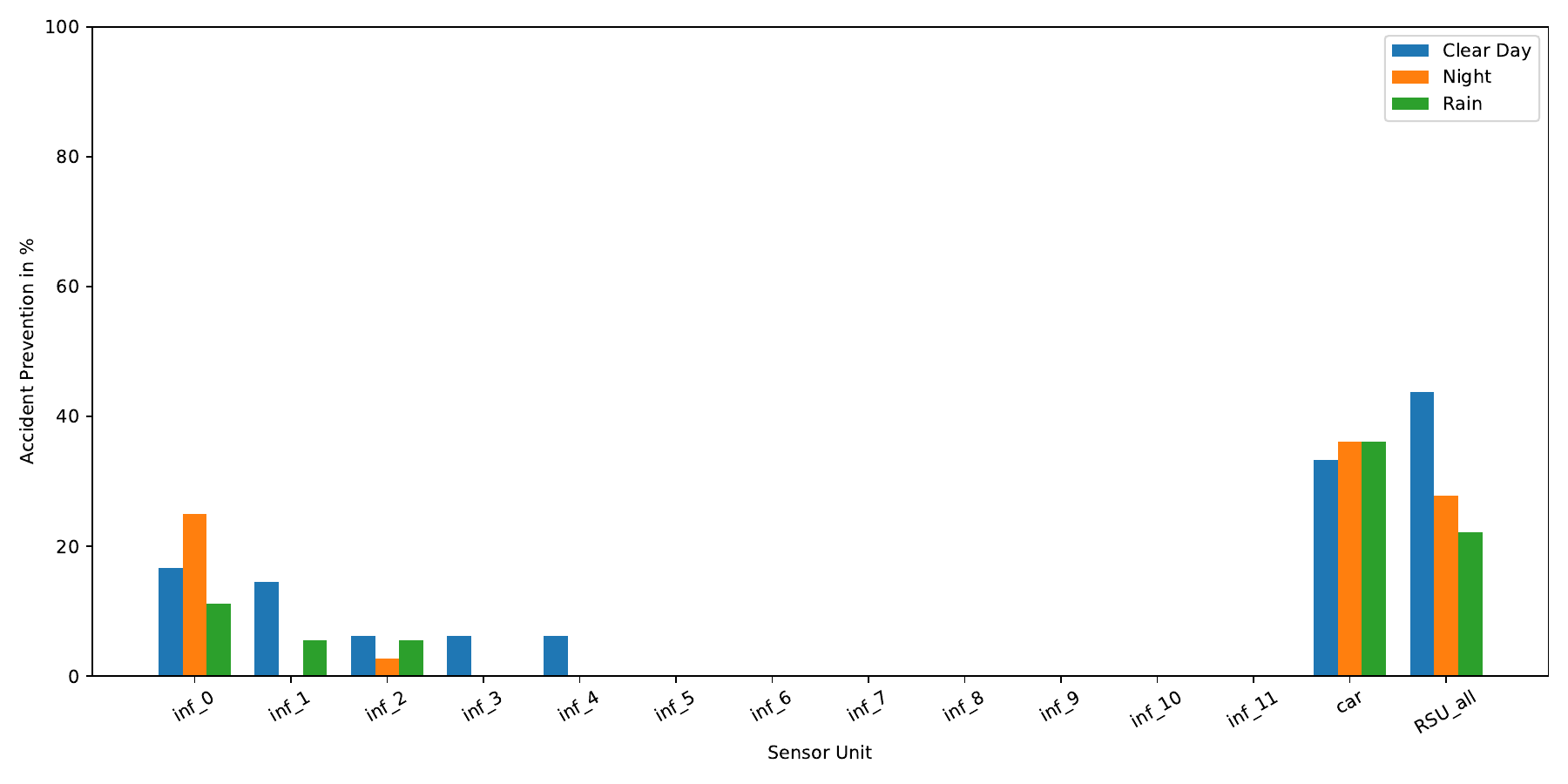}
        \caption{CBTAfo}
        \label{fig:CBTAfo_results}
    \end{subfigure}%
    \begin{subfigure}[t]{0.43\textwidth}
        \centering
        \includegraphics[width=\linewidth]{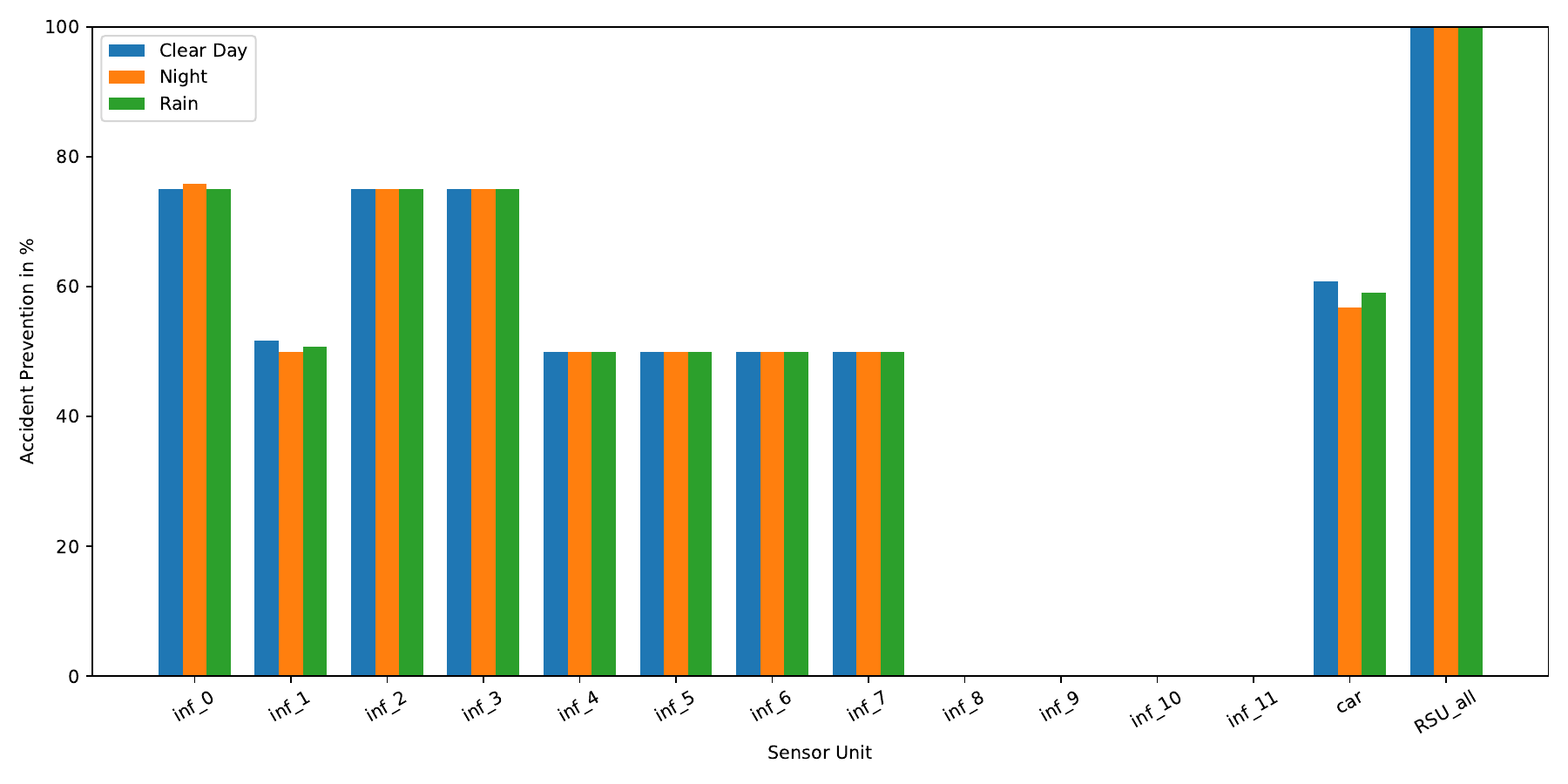}
        \caption{CPNCO}
        \label{fig:CPNCO_results}
    \end{subfigure}
    
    \begin{subfigure}[t]{0.43\textwidth}
        \centering
        \includegraphics[width=\linewidth]{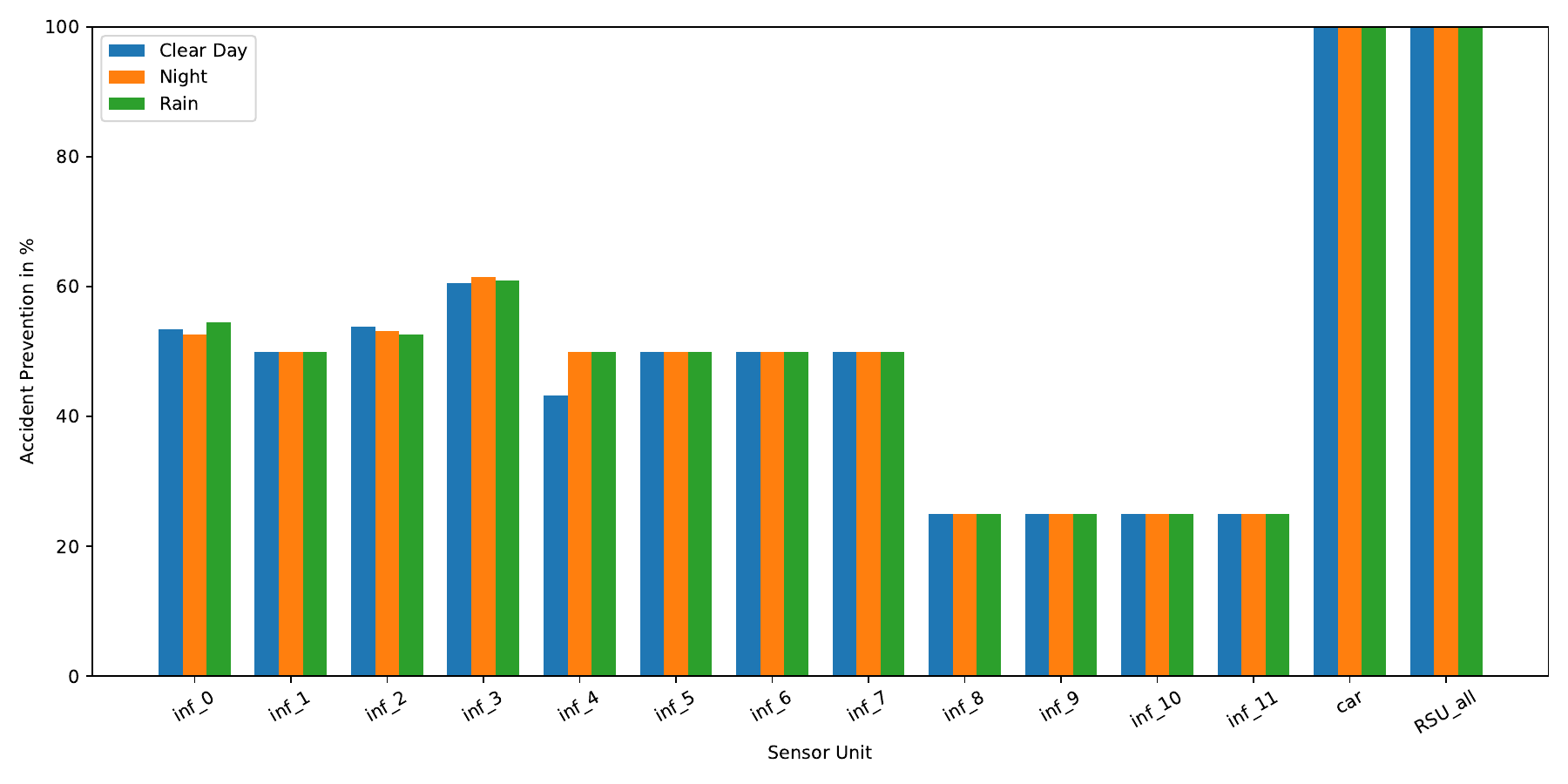}
        \caption{CPLA}
        \label{fig:CPLA_results}
    \end{subfigure}
    \begin{subfigure}[t]{0.43\textwidth}
        \centering
        \includegraphics[width=\linewidth]{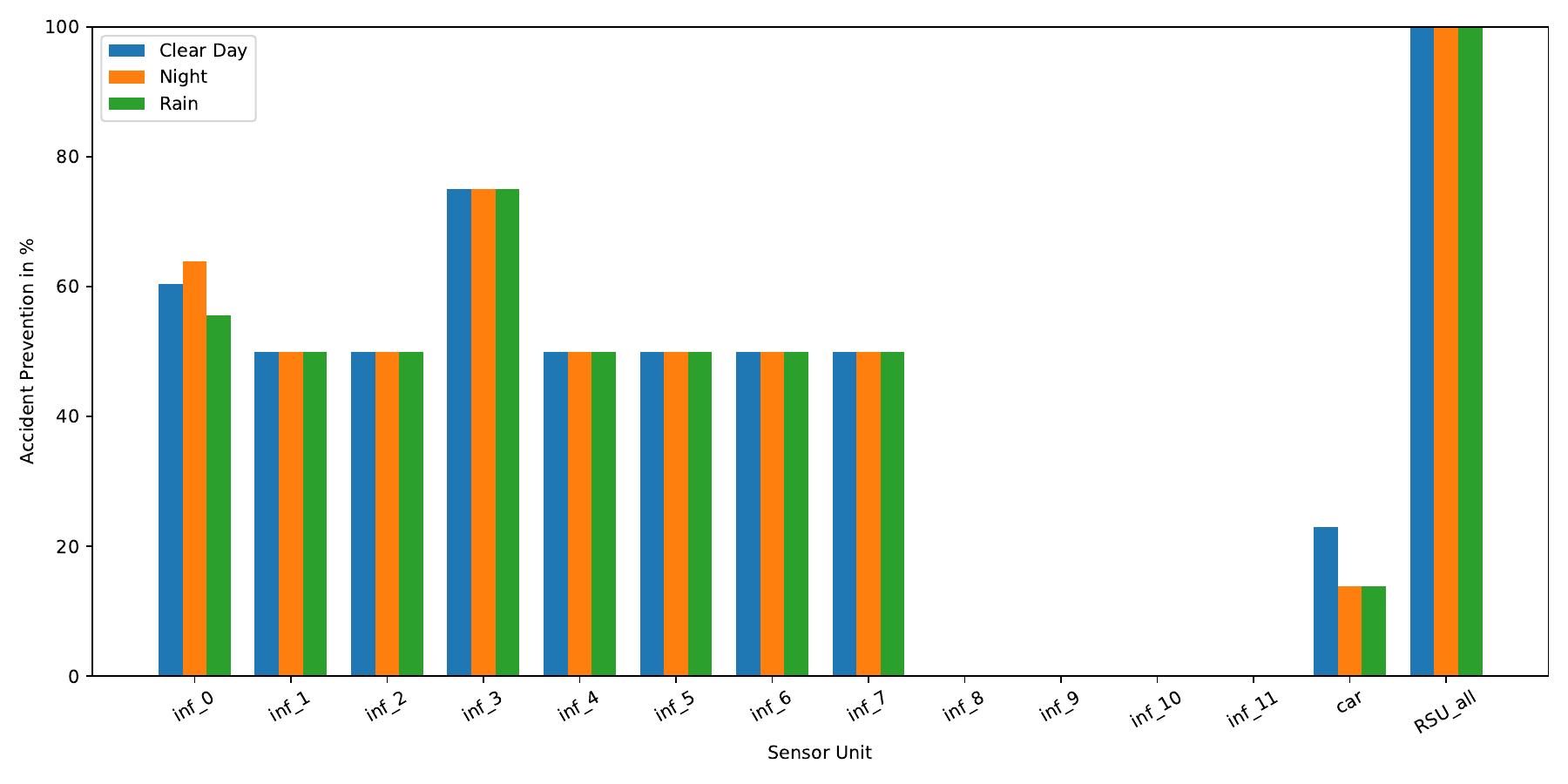}
        \caption{CPTAns}
        \label{fig:CPTAns_results}
    \end{subfigure}
    \caption{Results for the EuroNCAP VRU AEB scenarios (Protocol Version 4.5.1) included in the CarlaNCAP framework with clear weather (blue), at night (orange) and under rainy conditions (green).}
    \label{fig:carla_ncap_new_results}
\end{figure*}

The collision avoidance rate over all speeds for each scenario is shown in Fig.~\ref{fig:results_safety} and Fig.~\ref{fig:carla_ncap_new_results}.

For the CPNC-50 scenario the pedestrian could be detected by the VUT at \SI{13.97}{\percent} of the time which allows the VUT to avoid a collision in \SI{65.63}{\percent} of the scenarios.
\begin{figure}[H]
    \centering
    \includegraphics[width=\linewidth]{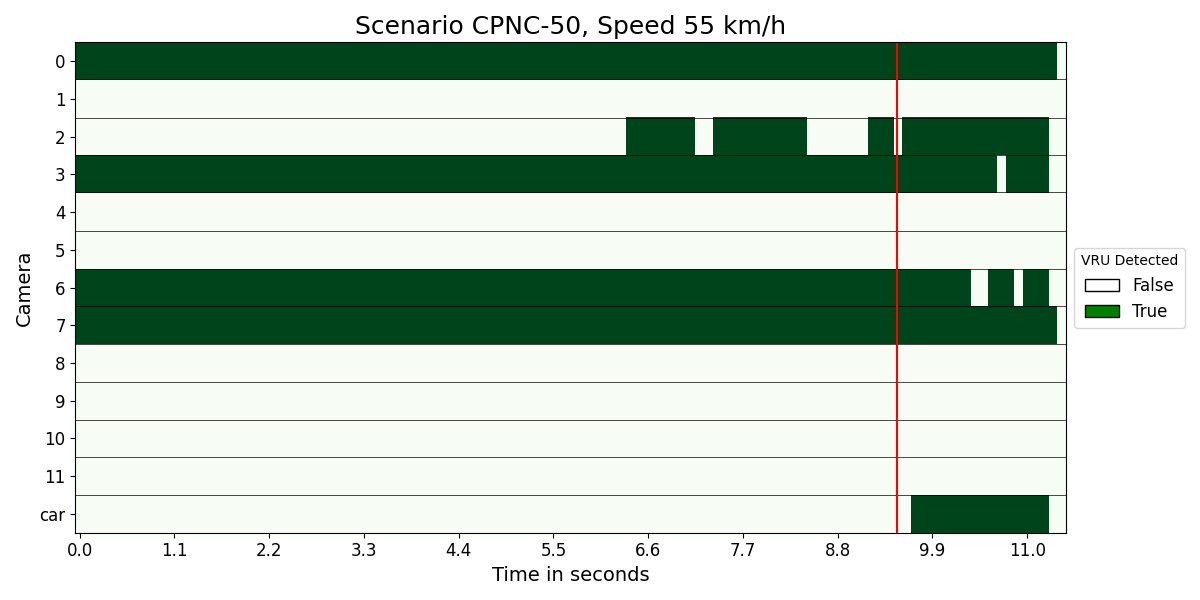}
    \caption{Heatmap for the detection on the CPNC-50 scenario with a VUT speed of \SI{55}{\kilo\metre\per\hour}. Green areas mark a correct detection. The red line represents the last possible time to start the emergency braking maneuver.}
    \label{fig:cpnc_heatmap}
\end{figure}

Considering the RSUs, in \SI{98.93}{\percent} of the frames, at least one RSU has detected the VRU. Some of the RSU cameras (8-11) were not 
able to detect the VRU for at least three consecutive frames, resulting in no safety enhancement.

However, the field of view of these cameras is in direction of the incoming roads, thus, the VRU is not in their FOV. For \textit{RSU2} and \textit{RSU3} a collision avoidance rate of \SI{100}{\percent} can be observed. Additionally, \textit{RSU0} reaches about \SI{97}{\percent}. A more detailed look into the results is provided in Fig.~\ref{fig:cpnc_heatmap}. Here, the detection over time for the CPNC-50 scenario with a VUT speed of \SI{55}{\kilo\metre\per\hour} is shown. It can be observed that the VUT is not capable to detect the VRU before the last possible time step to start the AEB maneuver. It can be observed that four RSUs in this scenario can correctly detect the VRU for nearly all frames. A fifth RSU (\textit{RSU2}) is also able to detect the VRU for some frames before the last possible time step to start the AEB maneuver.

\begin{figure}
    \centering
    \includegraphics[width=\linewidth]{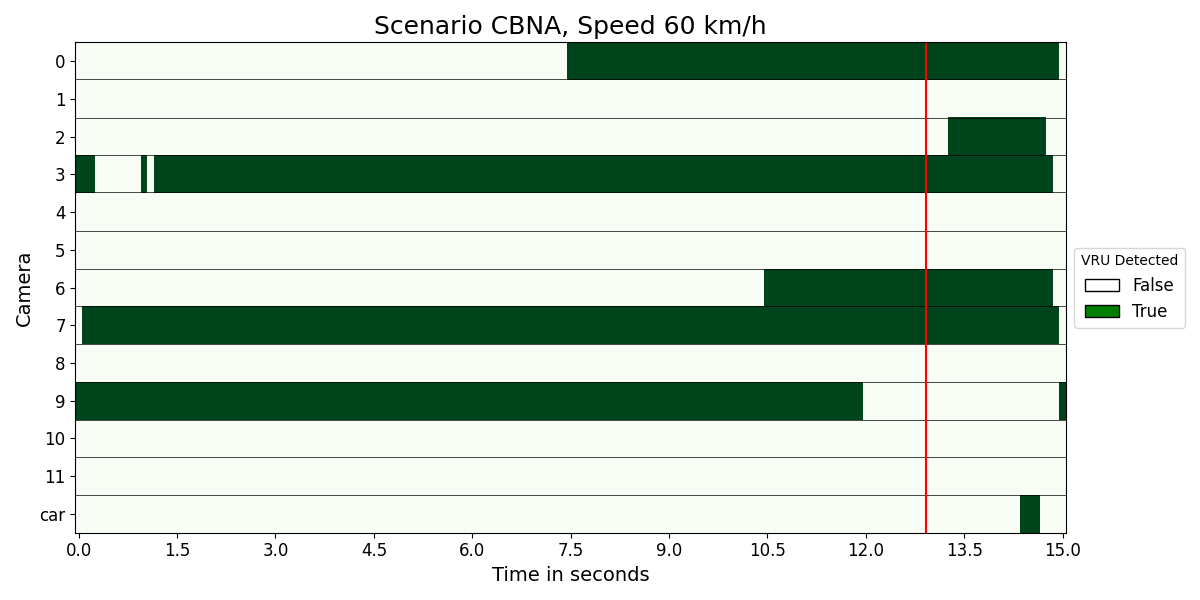}
    \caption{Heatmap for the detection on the CBNA scenario with a VUT speed of \SI{60}{\kilo\metre\per\hour}. Green areas mark a correct detection. The red line represents the last possible time to start the emergency braking maneuver.}
    \label{fig:cbna_heatmap}
\end{figure}

For the CBNA scenario where the cyclist enters the intersection from the right, occluded by a wall, for the VUT only a collision avoidance of about \SI{33}{\percent} could be achieved. Due to the occlusion, the VUT is only able to perceive the VRU at \SI{5.12}{\percent} of the frames, which are mostly the last frames shortly before the collision. As for the CPNC-50 scenario, \textit{RSU2} and \textit{RSU3} reach a collision avoidance rate of \SI{100}{\percent}. In addition different other RSUs were able to significantly increase the collision avoidance to about \SI{80}{\percent}. Overall, the VRU was detected by at least one RSU in \SI{99.43}{\percent} of the frames. An exemplary heatmap for the VUT speed of \SI{60}{\kilo\metre\per\hour} is shown in Fig.~\ref{fig:cbna_heatmap}. For the VUT, the VRU can be detected at about \SI{14.5}{\second} which is about \SI{1.5}{\second} after the last possible time step to avoid a collision. In this scenario, four RSUs were able to correctly detect the VRU which could avoid the collision.

\begin{figure}[b]
    \centering
    \includegraphics[width=\linewidth]{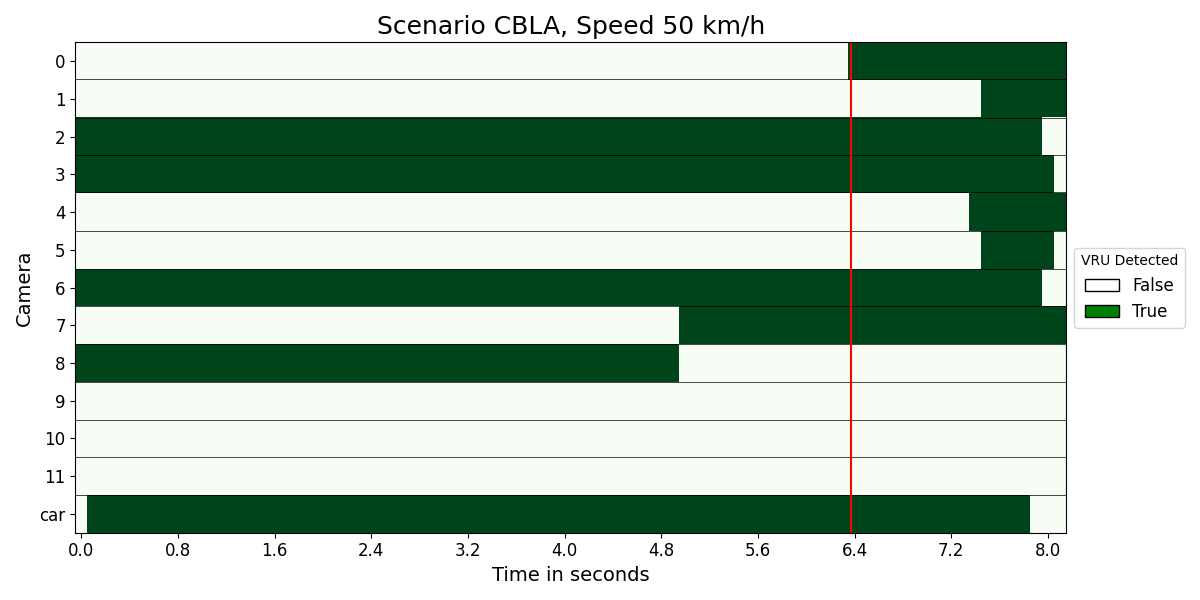}
    \caption{Heatmap for the detection on the CBLA scenario with a VUT speed of \SI{50}{\kilo\metre\per\hour}. Green areas mark a correct detection. The red line represents the last possible time to start the emergency braking maneuver.}
    \label{fig:cbla_heatmap}
\end{figure}

In the CBLA scenario the VUT was able to detect the VRU in \SI{89.20}{\percent} of the frames which results in a collision avoidance rate of \SI{100}{\percent}. However, in this scenario no occlusion is present, which makes it much easier to perceive the VRU driving in front of the VUT. For the RSUs, the VRU was detected in \SI{99.49}{\percent} of the frames. The collision avoidance heavily depends on their viewing direction and ranges between \SIrange{0}{100}{\percent}. This is also shown in Fig.~\ref{fig:cbla_heatmap}. 

%\vspace*{-2mm}

Additionally, it can be observed that even for a VUT speed of \SI{50}{\kilo\metre\per\hour} the VUT is capable of detecting the VRU for almost the complete scenario and could avoid a collision without the support of infrastructure-assisted CP. 
It can be observed that for over all scenarios and VUT speeds, the ratio of frames in which the VRU is detected correctly, ranges between 0.926 for CBLA at \SI{20}{\kilo\meter\per\hour} and 1.00 for various scenarios. Mostly, the result is above 0.99, which indicates that the comprehensive sensor setup used in our experiments allows for a VRU detection at nearly each time step. 
Over all scenarios the VUT collision avoidance ranges between \SIrange{33}{100}{\percent}, where the \SI{100}{\percent} only could be achieved in a scenario without any occlusion. Especially, for the CBNA scenario the necessity for a perception improvement is demonstrated. Here, two of the RSUs were able to increase the collision avoidance rate from about \SI{33}{\percent} to \SI{100}{\percent}. In addition it can be observed that \textit{RSU3} is able to achieve a collision avoidance rate of \SI{100}{\percent} over all scenarios. This demonstrates that even a small infrastructural sensor setups can significantly increase the collision avoidance rate and contribute to VRU safety.

For the additional scenarios, it can be observed that the accident prevention rates are significantly lower compared to the initial scenarios. This can primarily be attributed to the increased complexity of the scenarios. For the CBNA scenario, the accident prevention ratio (APR) ranges in \SIrange{0}{31} {\percent} in clear weather, with the best results achieved by the car and RSU4. Due to the high velocity of the bicycle and the short distance between the traffic participants, avoiding the collision becomes particularly challenging.

For the turning scenarios, CBTAno and CBTAfo, it can be observed that most RSUs are not capable of significantly increasing safety, while the car performs best among the individual sensor units. This behavior can be traced back to the current sensor layout, as the cameras are oriented toward the incoming roads and the inner area of the intersection, whereas the vehicle approaches from outside the main field of view and therefore remains largely in a blind spot. Consequently, the RSUs only achieve \SIrange{0}{25}{\percent}, while the car achieves \SI{33}{\percent}. Furthermore, accidents can only be prevented at comparatively low velocities. When considering the combination of all RSUs as a single sensing entity (RSU\_all), a significant increase in APR of up to approximately \SI{95}{\percent} is achieved for CBNA. For the other two scenarios, the improvement is less pronounced; however, the combined RSUs achieve performance comparable to that of the car.

Considering the pedestrian scenarios, similar results for CPNCO as for the CPCN-50 scenario of the initial work can be observed. While the car only achieves an APR of approximately \SI{60}{\percent}, the individual RSUs achieve up to \SI{75}{\percent}. For the CPLA scenario, the car performs best with an APR of \SI{100}{\percent}. Since the pedestrian walks on the road directly in front of the car, no major occlusions occur and the pedestrian can be perceived reliably by the vehicle sensors. In contrast, the performance of the RSUs, which achieve up to \SI{60}{\percent}, strongly depends on the current position of the VRU and whether it is located within the field of view of the sensors. For the turning scenario CPTAns, the RSUs again significantly increase the safety of the VRU, achieving an APR of up to \SI{75}{\percent}, corresponding to an improvement of approximately 50 percentage points compared to the car. When combining all RSUs (RSU\_all), a substantial increase can be observed across all conditions, resulting in an APR of \SI{100}{\percent}.

The low-lighting conditions lead to only a minor effect on the perception performance. As shown across the different scenarios, the night-time performance is generally comparable to the daytime performance, with deviations typically below 3 percentage points. This can be attributed to the scenario setup. In urban environments, various street lamps are installed along the sidewalks to support pedestrians in dark conditions. Moreover, the VUT operates with activated headlights, providing an additional source of illumination. Similarly, the effect of rain in these scenarios is mostly negligible, particularly for the pedestrian scenarios. For the bicycle scenarios, however, slight degradations can be observed. For example, in CBTAfo for RSU3 and RSU4, an APR of approximately \SI{6}{\percent} is achieved in clear weather, whereas under rainy conditions the APR decreases to \SI{0}{\percent}.

% \begin{table*}
%       \caption{Accuracy results for all VUT speeds and EuroNCAP test cases with clear weather and daytime}
%       \centering

% \renewcommand{\arraystretch}{1.05}
% %\resizebox{\textwidth}{!}{%
%     \begin{tabular}{cccccccccc} \toprule
%     &  \multicolumn{9}{c}{\textbf{Speed of Vehicle under Test}} \\
%     \cline{2-10}\\
%     \textbf{Test case} &\SI{20}{\kilo\meter\per\hour}&\SI{25}{\kilo\meter\per\hour}&\SI{30}{\kilo\meter\per\hour}&\SI{35}{\kilo\meter\per\hour}&\SI{40}{\kilo\meter\per\hour}&\SI{45}{\kilo\meter\per\hour}&\SI{50}{\kilo\meter\per\hour}&\SI{55}{\kilo\meter\per\hour}&\SI{60}{\kilo\meter\per\hour}\\ \midrule
%     CPNC-50& 0.952&0.998&0.996&0.990&0.990&0.996&0.990&0.992&0.990\\ 
%     CBNA& 0.954&0.999&1.000&0.998&0.998&0.998&0.992&1.000&1.000\\ 
%     CBLA& 0.926&1.000&1.000&1.000&1.000&1.000&1.000&1.000&1.000\\ 
%     \bottomrule
% \end{tabular}
% \label{tab:results}
% \end{table*}

%%%%%%%%%%%%%%%%%%%%%%%%%%%%%%%%%%%%%%%%%%%%%%%%%%%%%%%%%%%%%%%%%%%%%%%%%%%%%%%%
\section{CONCLUSION \& OUTLOOK}
\label{sec:conclusion}

In this paper, we presented a novel framework for evaluating the safety of vulnerable road users in urban environments with the novel synthetic CarlaNCAP dataset using standardized EuroNCAP scenarios. The dataset consists of 11,134 frames, and for each frame, image data and LiDAR point clouds from 12 roadside-units and one vehicle-under-test is provided, as well as 2D and 3D object information.

We demonstrated that vehicle-local perception could only avoid an accident in \SIrange{20}{66}{\percent} of the scenarios with occlusion of the VRU. Moreover, the conducted simulation study shows that already a small infrastructure-sensor setup with two to three cameras could increase the accident avoidance rate up to \SI{100}{\percent}.

For further research, possible extensions of the EuroNCAP test protocol will be included. Moreover, the evaluation will be extended to further environmental conditions such as fog and LiDAR-based object detection. In addition to roadside sensors, the VRU awareness message (VAM) which are broadcasted by mobile phones or smart bikes is a promising approach and could be included in the study to extend the information basis and enhance the VRU safety. 
%\newpage
%%%%%%%%%%%%%%%%%%%%%%%%%%%%%%%%%%%%%%%%%%%%%%%%%%%%%%%%%%%%%%%%%%%%%%%%%%%%%%%%

%\addtolength{\textheight}{-6.5cm}

%%%%%%%%%%%%%%%%%%%%%%%%%%%%%%%%%%%%%%%%%%%%%%%%%%%%%%%%%%%%%%%%%%%%%%%%%%%%%%%%

\bibliographystyle{IEEEtran} % use IEEEtran.bst style
\bibliography{literature.bib}

\end{document}